\documentclass[10pt,twocolumn,letterpaper]{article}

\usepackage[pagenumbers]{config/cvpr}

\newcommand{\ourTitle}{\modelname: Generating 4D Whole-Body Motion for Hand-Object Grasping}

\newcommand{\modelnameCOLOR}{black}
\newcommand{\ourTitleSUPMAT}{\ourTitle\\ \emph{*Supplemental Material*}}
\newcommand{\modelname}{\textcolor{\modelnameCOLOR}{\mbox{GOAL}}\xspace}
\newcommand{\modelnameLong}{\textcolor{\modelnameCOLOR}{Generating Object-interActing whoLe-Body motions}}

\usepackage[toc,page]{appendix}

\newcommand{\video}{\textcolor{\modelnameCOLOR}{\textbf{video}}\xspace}
\newcommand{\supmat}{\textcolor{\modelnameCOLOR}{{Sup.~Mat.}}\xspace}

\usepackage{graphicx}
\usepackage{amsmath}
\usepackage{amssymb}
\usepackage{booktabs}
\usepackage{bm}
\usepackage{multicol}

\usepackage{balance}

\usepackage[pagebackref,breaklinks,colorlinks]{hyperref}

\usepackage{cuted}
\usepackage{enumitem}

\usepackage{multirow}
\usepackage{booktabs}
\usepackage{colortbl}

\usepackage[dvipsnames]{xcolor}  %

\newcommand{\change}[1]{{\color{black} #1}}
\newcommand{\TODO}[1]{\textcolor{black}{#1}}

\newcommand{\colorRef}[1]{\textcolor{black}{#1}} %
\usepackage[capitalize]{cleveref}
\crefname{figure}{\colorRef{Fig.}}{\colorRef{Figs.}}
\Crefname{figure}{\colorRef{Figure}}{\colorRef{Figures}}
\crefname{section}{\colorRef{Sec.}}{\colorRef{Secs.}}
\Crefname{section}{\colorRef{Section}}{\colorRef{Sections}}
\Crefname{table}{\colorRef{Table}}{\colorRef{Tables}}
\crefname{table}{\colorRef{Tab.}}{\colorRef{Tabs.}}

\newcommand{\mesh}{M}
\newcommand{\shape}{\bm{\beta}}

\newcommand{\pose}{\bm{\theta}}
\newcommand{\expression}{\bm{\psi}}

\newcommand{\nvertices}{10,475\xspace}

\renewcommand{\etal}{et al.\xspace}
\renewcommand{\ie}{i.e.\xspace}
\renewcommand{\eg}{e.g.\xspace}

\newcommand{\gnet}{\mbox{GNet}\xspace}
\newcommand{\mnet}{\mbox{MNet}\xspace}

\newcommand{\Gnet}{\gnet}
\newcommand{\Mnet}{\mnet}

\newcommand{\smplx}{\mbox{SMPL-X}\xspace}
\newcommand{\smplX}{\smplx}
\newcommand{\mano}{\mbox{MANO}\xspace}

\newcommand{\grab}{GRAB\xspace}

\newcommand{\grabnet}{\mbox{GrabNet}\xspace}

\newcommand{\mocap}{\mbox{MoCap}\xspace}

\newcommand{\threeD}{\xspace{3D}\xspace}

\newcommand{\bps}{BPS\xspace}

\newcommand{\goal}{{g}}

\newcommand{\cVAE}{\mbox{cVAE}\xspace}

\newcommand{\groundtruth}{{ground-truth}\xspace}

\DeclareSymbolFont{matha}{OML}{txmi}{m}{it}%
\DeclareMathSymbol{\varv}{\mathord}{matha}{118}

\newcommand{\qheading}[1]{\noindent\textbf{#1}}

\newcommand{\subject}[0]{}

\newcommand{\object}[0]{\mathtt{o}}
\newcommand{\head}[0]{\mathtt{h}}
\newcommand{\foot}[0]{\mathtt{f}}
\newcommand{\reals}[0]{\mathbb{R}}
\newcommand{\normabs}[1]{\lVert#1\rVert_1}
\newcommand{\normmse}[1]{\lVert#1\rVert_2}

\newcommand{\loss}[0]{\mathcal{L}}
\newcommand{\energy}[0]{\bm{E}}

\newcommand{\encoder}[0]{\mathcal{E}}

\newcommand{\graspcode}[0]{\bm{z}_g}

\newcommand{\transl}[0]{\bm{t}}
\newcommand{\translobject}[0]{\transl^{\object}}
\newcommand{\basis}[0]{\bm{b}}
\newcommand{\basisobject}[0]{\basis^{\object}}

\newcommand{\vel}[1]{\dot{#1}}
\newcommand{\vertsvel}[0]{\vel{\verts}}

\newcommand{\smplxparams}[0]{\bm{\Theta}^{\subject}}

\newcommand{\smplxparamsGT}[0]{\hat{\smplxparams}}

\newcommand{\smplxparamssubject}[0]{\smplxparams}

\newcommand{\translsubject}[0]{\transl^{\subject}}
\newcommand{\shapesubject}[0]{\shape^{\subject}}
\newcommand{\vertssubject}[0]{\verts^{\subject}}
\newcommand{\vertssubjectGT}[0]{\hat{\verts}^{\subject}}
\newcommand{\headorient}[0]{\bm{h}^{\subject}}
\newcommand{\headorientGT}[0]{\hat{\bm{h}}^{\subject}}
\newcommand{\offsetssubject}[0]{\offsets^{b \rightarrow o}}
\newcommand{\posesubject}[0]{\pose^{\subject}}

\newcommand{\vtov}{\mbox{V2V}\xspace}

\newcommand{\verts}[0]{\bm{v}}
\newcommand{\vertsobject}[0]{\verts^{\object}}
\newcommand{\righthand}[0]{h}
\newcommand{\vertsrh}[0]{\verts^{\righthand}}
\newcommand{\vertsrhGT}[0]{\hat{\verts}^{\righthand}}

\newcommand{\offsets}[0]{\bm{d}}
\newcommand{\offsetsGT}[0]{\hat{\bm{d}}}

\DeclareMathOperator*{\argmin}{arg\,min}

\usepackage{lipsum}
 
\newcommand{\websiteURL}{\mbox{\url{https://goal.is.tuebingen.mpg.de}}}

\usepackage[normalem]{ulem} %

\usepackage{acronym}
\acrodef{amt}[AMT]{Amazon Mechanic Turk}    %

\usepackage{hhline}

\begin{document}

\title{\ourTitle}

\author{
Omid Taheri \quad Vasileios Choutas \quad Michael J. Black \quad Dimitrios Tzionas \\
Max Planck Institute for Intelligent Systems, T{\"u}bingen, Germany
\\
{\tt\small \{otaheri, vchoutas, black, dtzionas\}@tue.mpg.de
}
}

\twocolumn[
{
    \renewcommand\twocolumn[1][]{#1}
    \vspace{-1.em}
    \maketitle
    \centering
    \vspace{-0.7em}
    \begin{minipage}{1.00\textwidth}
    \centering			%
	\includegraphics[width=1.00 \linewidth]{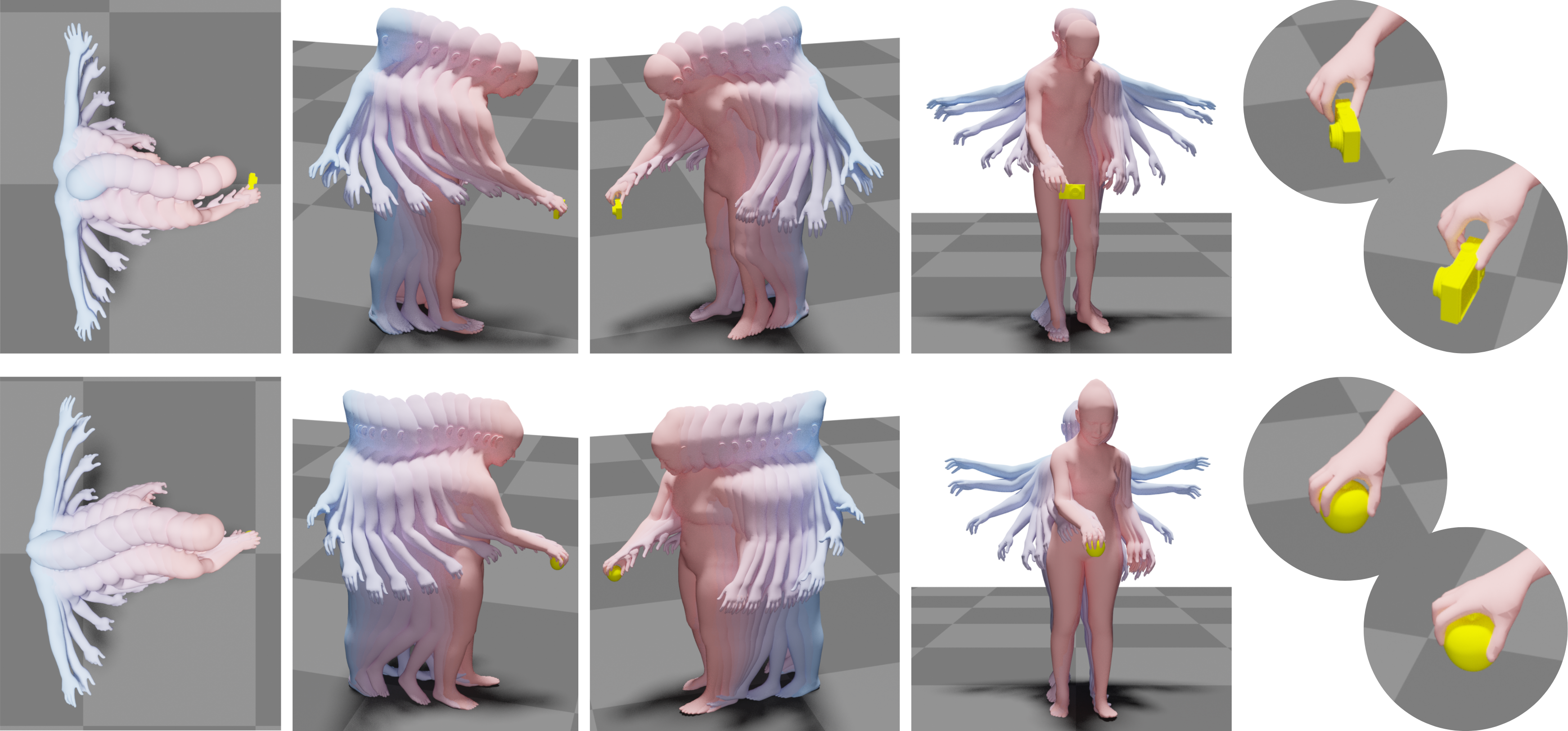}
    \end{minipage}
    \vspace{-0.8 em}
    \captionof{figure}{
                        \modelname generates whole-body motions for approaching and grasping an unseen \threeD object. 
                        The figure shows {\em generated motions} for $2$ people (top, bottom), each grasping a different novel object. 
                        For each sequence we show $4$ different views (left to right), as well as 
                        zoomed-in ``circle'' snapshots of the final grasp. 
                        \modelname is the first method to generate such a natural motion and grasp for the full body. 
    }\label{fig:teaser}
    \vspace*{+02.em}
}
]

\begin{abstract}
\vspace{-1.0 em}
Generating digital humans that move realistically has many applications and is widely studied, but existing methods focus on the major limbs of the body, ignoring the hands and head.
Hands have been separately studied but the focus has been on generating realistic \TODO{static} grasps of  objects. %
To synthesize virtual characters that \textsl{interact} with the world, we need to generate full-body motions and realistic hand grasps simultaneously.
Both sub-problems are challenging on their own and, together, the state-space of poses is significantly larger, the scales of hand and body motions differ, and the whole-body posture and the hand grasp must agree, satisfy physical constraints, and be plausible.
Additionally, the head is involved because the avatar must look at the object to interact with it. 
For the first time, we address the problem of generating full-body, hand and head motions of an avatar grasping an unknown object.
As input, our method, called \modelname, takes a \threeD object, its position, and a starting \threeD body pose and shape. 
\modelname outputs a sequence of whole-body poses using two novel networks. 
First, \textsl{\Gnet} generates a \textsl{goal} whole-body grasp with a realistic body, head, arm, and hand pose, as well as hand-object contact.
Second, \textsl{\Mnet} generates the \textsl{motion} between the starting and goal pose. 
This is challenging, as it requires the avatar to walk towards the object with foot-ground contact, orient the head towards it, reach out, and grasp it with a realistic hand pose and hand-object contact.
To achieve this the networks exploit a representation that combines \smplx body parameters and {\threeD  vertex offsets}.
We train and evaluate \modelname, both qualitatively and quantitatively, on the  \grab dataset. 
Results show that \modelname generalizes well to unseen objects, outperforming baselines. 
A perceptual study shows that \modelname's generated motions approach the realism of \grab's ground truth. 
\modelname takes a step towards synthesizing realistic full-body object grasping. 
Our models and code are available for research purposes at \websiteURL. 
\end{abstract}

\section{Introduction}	\label{sec:intro}

Virtual humans are important for movies, games, AR/VR and the metaverse. 
Not only do they need to look realistic, %
but also move and \textit{interact} realistically. 
Most work   on human motion generation %
            has focused only on bodies, without the head and hands. 
            Often, these bodies are considered in ``isolation'', with no scene or object context. 
Other work  focuses on bodies interacting with scenes, %
            but ignores the hands.
Similarly, work on generating hand grasps %
            often ignores the body. 
We argue that these are all just parts of the problem. 
What we really need, instead, is to generate motion of \emph{full-body} avatars \emph{grasping} objects, by jointly considering the body, head, hands, and the object. 
We address this here for the first time. 

\begin{figure}
    \centering
    \includegraphics[width=1.0\columnwidth]{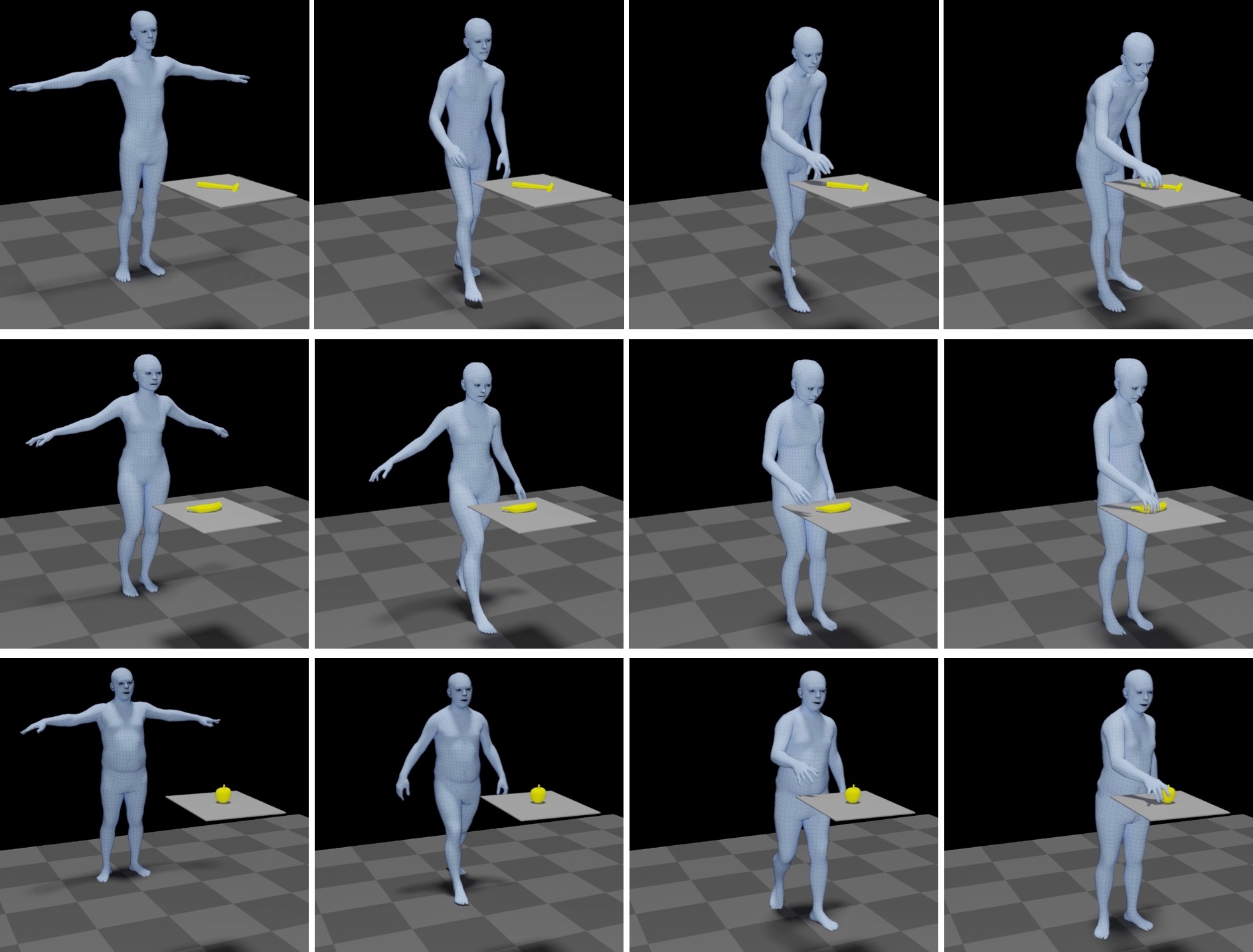}
    \vspace{-1.5 em}
    \caption{
                        Grasping an object involves several motions. 
                        We walk towards the object with our feet contacting the floor, 
                        we orient our head to look at the object, 
                        we lean our torso, extend our arms, 
                        and pose our hand to contact and grasp the object. 
                        The depicted examples use motions captured in the \grab dataset \cite{GRAB:2020}. 
    }
    \label{fig:motivation}
    \vspace{-0.5 em}
\end{figure}

The problem is challenging and multifaceted. 
Think of how we grasp objects in real life (see \cref{fig:motivation}); 
we walk towards the object with our feet contacting the floor, 
we orient our head to look at the object, 
lean our torso and extend our arms to reach it, and  
dexterously pose our hands to establish fine contact and grasp it. 
Humans are able to gracefully execute these steps, yet, these are challenging and involve motion planning, motor control, and spatial awareness. 
Some of these steps have been studied separately, but we cannot simply combine the partial solutions since the entire action must be {\em coordinated}.
This is challenging because: (1) full bodies have a much higher-dimensional state space than bodies or hands alone; (2)
the body and hands have very different sizes, motion scales and level of dexterity; (3) the body, head, and hands must move in a coordinated fashion.
Currently, there are no automatic tools to generate such coordinated full-body grasping motions.

\begin{figure*}%
    \centering
    \includegraphics[width=1.0\textwidth]{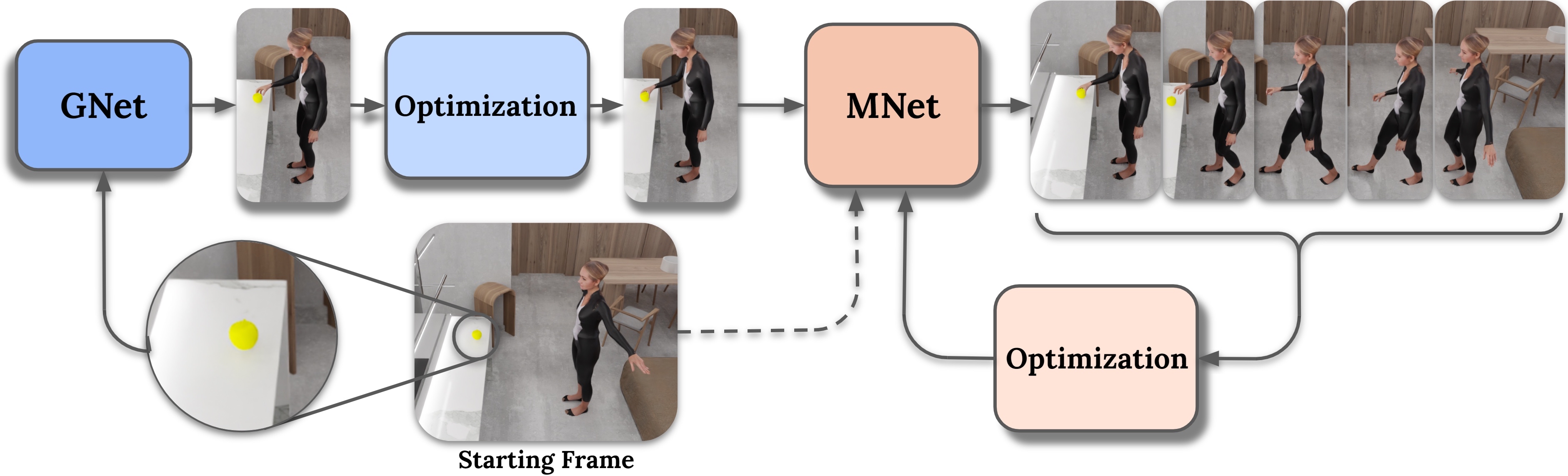}
    \caption{
                Overview of \modelname.  
                There are two main stages: 
                (1)     \Gnet takes as input the object and its location, and generates a ``goal'' whole-body grasping pose. 
                        The output pose is refined with optimization post processing to look more realistic and physically plausible. 
                (2)     \Mnet takes as input a starting pose and the generated ``goal'' pose for the human, and 
                        generates the motion in between as a sequence of poses in an auto-regressive fashion. 
                        The output poses are refined with optimization post processing 
                        to better ``reach'' the ``goal'' pose. 
    }
    \label{fig:goal_setup}
\end{figure*}

We address this with \emph{\modelname}, which stands for \emph{\modelnameLong}. 
\modelname generates whole-body avatar motion for grasping an unknown object, by jointly considering the body, head, hands, and the object. 
\modelname takes three \emph{inputs}:
(1)     a \threeD object, 
(2)     its position and orientation, and 
(3)     a ``starting'' \threeD body pose and shape, positioned near 
        the object and roughly oriented towards it. 
As output, \modelname generates a sequence of 3D body poses from the starting pose through to an object grasp.
To do so, \modelname uses two novel networks (for an overview see \cref{fig:goal_setup}): 
\mbox{(1)   First, \Gnet generates} a ``goal'' whole-body  grasp, with  a  realistic  body pose, head pose, arm pose, and hand pose, 
            as well as realistic finger-object and foot-ground contact. 
            \Gnet is formulated as a conditional variational auto-encoder (\cVAE), thus, it learns a distribution over grasping poses, and 
            can generate a variety of ``goal'' grasps. 
\mbox{(2)   Then, \Mnet inpaints} the motion between the ``starting'' and ``goal'' poses, by generating a sequence of whole-body poses in an auto-regressive fashion. 
This is challenging because
the avatar needs to (see \cref{fig:teaser}) 
walk by taking a number of steps proportional to the distance to the object, while having natural foot-floor contact without ``skating'', and 
continuously orient the head to look at the object. 
Then, when it is near the object, it needs to slow down, stop walking, 
lean the torso, 
extend the arms to reach the object.
It must also pose the hand to 
contact the object and grasp it. 
All body parts need to move gracefully and in full coordination, so that the motion looks natural.

Achieving this level of realism requires technical novelties.
\modelname draws inspiration by recent work \cite{zhang2021mojo,zanfir2021thundr,Loper:SIGASIA:2014}, 
but goes beyond this to uniquely infer both \smplX \cite{smplifyPP} parameters and \threeD offsets. %
\Gnet infers \threeD hand-to-object vertex offsets to give spatial awareness and guide object grasping. 
\Mnet infers \threeD \smplX vertex offsets to guide \smplX deformation from the previous to the current frame. 
These offsets lie in \threeD Euclidean space, thus, they can be more accurately inferred than \smplX parameters, and are used in an offline optimization scheme to refine \smplX poses. 
We train \Gnet and \Mnet on the \grab~\cite{GRAB:2020} dataset, which contains whole-body \smplX humans grasping objects.

We evaluate \modelname, both quantitatively and qualitatively, on withheld parts of the \grab dataset. 
Specifically, we withhold 5 objects %
for testing.
Results show that \modelname generalizes well and produces natural motions for full-body walking and object grasping; see \cref{fig:teaser}. %
Quantitative evaluation shows that \modelname outperforms baselines, and ablation studies show a positive contribution of all major components. 
A perceptual study, verifies the above, while showing that \modelname's generated motions achieve a level of realism comparable to \grab's \groundtruth motions.

To conclude, \modelname takes a step towards automatic whole-body grasp motion generation for realistic avatars. 
Models and code will be available for research purposes.

\section{Related Work}	\label{sec:related}

\textbf{Motion generation for bodies ``in isolation'':}
Research on human motion generation has a long history \cite{badler1993simulatingHumans,brand2000styleMachines,wang2008gaussianProccMotion}. 
However, even recent methods \cite{petrovich2021actor,Mao_Historyrepeats_2020_ECCV,yuan2020dlow,zhang2021mojo}, mostly study the body ``in isolation''; \ie, with no scene context. 
Most methods generate the motion of \threeD skeletons \cite{martinez2017motionPrediction,Mao_Historyrepeats_2020_ECCV,Gopalakrishnan_2019_CVPR,yuan2020dlow,holden2016deep,mao2019learningTraj}, while 
others \cite{zhang2021mojo,petrovich2021actor,chuan2020action2motion} 
generate the motion of %
a human model like SMPL \cite{SMPL:2015}. 
Typically, $1$-$2$ seconds of motion synthesis is referred to as ``long term". 
Early deep-learning methods employ RNNs \cite{Martinez_2017_ICCV,Gopalakrishnan_2019_CVPR,Fragkiadaki_2015_ICCV}, however, 
they struggle with discontinuities between the observed and predicted poses, and with long-range spatial relations across time. 
Other methods account for these with phase-functioned feed-forward neural networks \cite{starke2019neuralStateMachine,holden2017phase}, \ie by conditioning the network weights on phase.  However, these focus on cyclic motions. %
More recent methods \cite{li_dance_2021_arxiv,Mao_Historyrepeats_2020_ECCV,Tang_ijcai2018,petrovich2021actor} adopt an attention \cite{Vaswani_attention_2017_nips} mechanism. %

\textbf{Motion generation for bodies in \threeD scenes:}
Most early methods extend \mocap databases with point annotations for foot and hand contact \cite{gleicher1998retargetting,lee2002interactive,lee2006motionPatches,kapadia2016precision}. Then, they fit motion to  contacts with optimization and space-time constraints for \threeD body motion re-targeting \cite{gleicher1998retargetting}, and animating bodies that move in \threeD terrains \cite{lee2002interactive,lee2006motionPatches,kapadia2016precision}. %
 
To avoid big \mocap datasets, some methods use deep reinforcement learning (RL) for 
body-scene  \cite{chao2019learning2sit,peng2018deepmimic,peng2016terrain} or 
hand-object \cite{GHernando2020physicsDext,chen2021systemHOI} interactions. 
These methods show promising results for navigating terrains with varying height and gaps \cite{peng2018deepmimic,peng2016terrain}, sitting on chairs \cite{chao2019learning2sit,Starke:ToG:2019}, %
using a hammer and opening a door \cite{GHernando2020physicsDext}, and for in-hand object re-orientation \cite{chen2021systemHOI}. 
Generalization to new bodies, object geometry, and interaction types remains a challenge.

Others follow a \threeD geometric approach. %
Pirk      \etal \cite{pirk2017understanding}    place virtual sensors on objects to sense the flow of points sampled on an agent interacting with these, and build functional object descriptors. 
Al-Asqhar \etal \cite{alAsqhar2013relationship} re-target body motion by encoding human joints \wrt fixed points sampled on a scene. %
Ho        \etal \cite{ho2010spatial} use body and object vertices to compute per-frame ``interaction meshes'', and minimize their Laplacian deformation to re-target body motion.
These pure geometric methods are not robust to real-world noise.

In contrast, we fall in the category of data-driven methods. 
Corona    \etal \cite{corona2020context}    generate the context-aware motion of a human skeleton interacting with objects, where ``context'' is encoded as a directed graph connecting person and object nodes. 
More relevant are methods for generating motion  between a ``start'' and a ``goal'' pose in a \threeD scene.
Hassan \etal~\cite{hassan2021samp} estimate a ``goal'' position and interaction direction on an object, plan a \threeD path  from a start body pose to this, and finally generate a sequence of body poses with an auto-regressive \cVAE for walking and interacting, \eg, sitting on a chair.
Wang \etal~\cite{wang2021synthesizingLongTerm} 	%
first estimate several ``sub-goal'' positions and bodies, divide these into 
short start/end pairs to synthesize short-term motions, and finally stitch these together in a long motion with an optimization process.

\textbf{Motion generation for hands:}
ElKoura1 \etal~\cite{elkoura2003handrix} estimate physically plausible hand poses for playing musical instruments, using a low dimensional pose space, with a data-driven approach.
Pollard \etal \cite{pollard2005physically} use \mocap to learn  a controller for physically-based grasping. 
Kry \etal \cite{kry2006interaction} capture hand \mocap and forces with sensors on objects, and use these to build ``interaction trajectories'', and synthesize and re-target motions with physics simulation. 
More related to us, Lie \etal~\cite{liu2012synthesisHandonly} take as input \mocap data of body and object motion, and add the missing hand motion to the body, by first searching for feasible contact point trajectories, and then generating smooth hand motion with space-time optimization that satisfies the estimated contacts.

\textbf{Pose generation for bodies in \threeD scenes:}
Early methods use either contact annotations \cite{lin2012sketching} or detections \cite{shape2pose2014} on \threeD objects, and fit \threeD skeletons to these. 
Other methods use physics simulation to reason about contacts and sitting comfort \cite{kang2014environment,leimer2020pose,zheng2015ergonomics}. 
Focusing on rooms instead of single objects, 
Grabner   \etal \cite{grabner2011chair}     predict all areas on a \threeD scene mesh where a \threeD human mesh can sit, using proximity and intersection metrics. %
Recent methods \cite{zhang2020generating,PLACE:3DV:2020,hassan2021posa} use deep learning to generate static humans interacting with a scene. 
Zhang 		\etal 	\cite{zhang2020generating} 	learn a \cVAE to generate  \smplX~\cite{smplifyPP} poses, conditioned on an input depth image and semantic segmentation of the scene. 
Zhang 		\etal 	\cite{PLACE:3DV:2020} use an explicit scene-centric representation of interaction, while 
Hassan \etal~\cite{hassan2021posa} use a human-centric representation.

\textbf{Pose generation for hand-object grasps:}
Taheri 	\etal~\cite{GRAB:2020} predict \mano~\cite{romero2017embodied} hand grasps for unseen \threeD object meshes, by first predicting a rough hand grasp, and then refining it with distance and contact metrics. 
Grady   \etal~\cite{grady2021contactOpt} refine grasps by first estimating contacts on both the hand and the object, and then refining the hand with optimization to satisfy the inferred contacts.

\textbf{Motion for full-body interactions:}
People use their body and hands together for interacting with the world. %
Hsiao et al. \cite{hsiao2006wholeBodyGrasp} build a database of whole-body grasps with a human operating an avatar, and perform imitation learning. 
Borras et al. \cite{asfour2015wholeTaxonomy} capture whole-body \mocap data \cite{kitMocap2015} of people interacting with scene objects and handheld objects, using a humanoid model, and define a pose taxonomy. 
Taheri \etal~\cite{GRAB:2020} capture whole-body \smplX~\cite{smplifyPP} interactions with handheld objects, but learn a \cVAE that generates only static grasping hands, due to the task complexity.
Merel \etal~\cite{merel2020catchCarry} use deep RL and human \mocap demonstrations to learn a vision-guided neural controller for %
picking up and carrying boxes, or catching/throwing a ball. 

\textbf{Summary:}
The community has focused on parts of the problem (either the body or the hands) or used unrealistic bodies.
\modelname learns to generate full-body \smplX motions, from walking to approach an object up to grasping it, given only a \threeD object and a starting human pose.

\section{Method}				\label{sec:method}

An overview of our method, \modelname, is shown in \cref{fig:goal_setup}. 
\modelname takes three \emph{inputs}, namely:  
(1)     a \threeD object, 
(2)     its position and orientation,  and 
(3)     a ``starting'' \threeD body pose and shape, 
positioned near the object (roughly $0.5-1.5$ m) and oriented towards it (roughly $\pm 10^{\circ}$). 
Then, as \emph{output}, \modelname generates \smplX motion with two main networks: 
(1)  \Gnet synthesizes a ``goal'' \smplX mesh that grasps the \threeD object 
            with a realistic body pose and hand-object contact;
(2)  \Mnet ``inpaints'' the motion from the starting to the ``goal'' frame, 
            by generating a sequence of ``moving'' \smplX bodies in an auto-regressive way. 
Without loss of generality, we model right-handed grasps.

\subsection{Human Model}		\label{sec:method_HumanModel}

We use the \smplX~\cite{smplifyPP} statistical \threeD whole-body model, which jointly captures the body, head, face and hands. 
\smplX is a differentiable function that takes as input shape, $\shape$, pose, $\pose$, and expression, $\expression$, parameters and then outputs a \threeD mesh, $\mesh$, with $\nvertices$ vertices, $V$, and $20,908$ triangles, $F$.
The shape vector $\shape \in \mathbb{R}^{20}$ contains coefficients of a low-dimensional space, created via PCA on \threeD meshes of roughly $4,000$ different people \cite{CAESAR}. 
The vertices are posed with linear blend skinning with a rigged skeleton,
$\mathcal{J} \in \mathbb{R}^{55 \times 3}$, that is learned from data.
Let $\smplxparams = \{\pose, \transl \}$ be the set of all \smplX parameters we will predict,
where 
$\posesubject   \in \reals^{55\times6}$ \cite{Zhou_2019_CVPR}, $\translsubject \in \reals^{3}$.
In the following, instead of using all the body vertices, we sample  $400$ vertices on body areas that are important for interactions, guided by the heatmaps of \grab~\cite{GRAB:2020}.

\subsection{Interaction-Aware Attention}    \label{sec:interactionAware}

Two common representations for body-object interaction are vertex-to-vertex distances between meshes and %
contact maps on meshes. 
However, the former carries information that is irrelevant to the interaction (\eg, vertices far away from the object), while the latter is too compact 
and carries no information about \threeD proximity before/after contact. 

\begin{figure}[t]
	\centering			%
	\includegraphics[trim=000mm 000mm 000mm 030mm, clip=false, width=1.00 \linewidth]{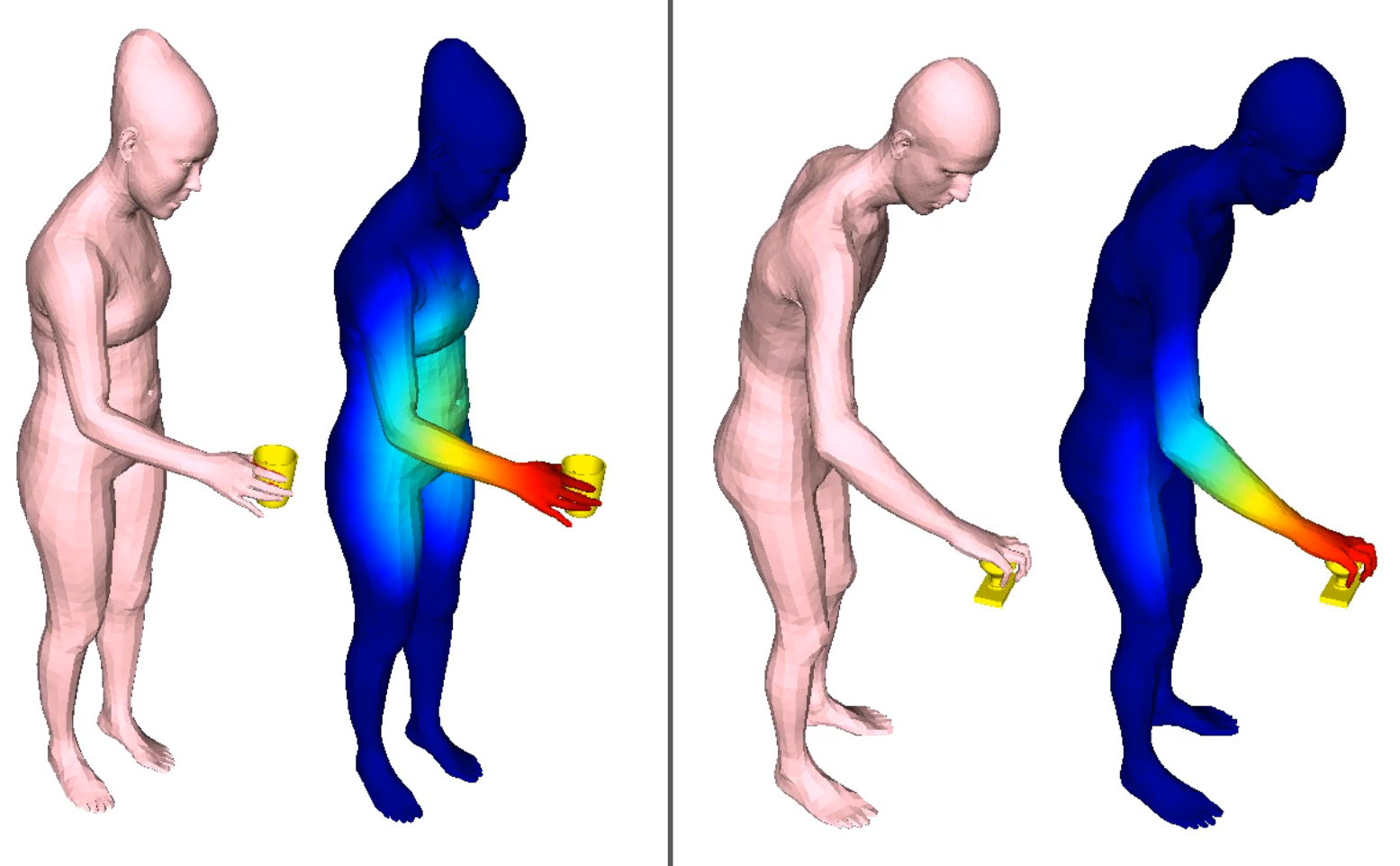}
	\caption{
	            Visualization of the ``interaction-aware'' attention for body-to-object vertex distances (\cref{sec:interactionAware}).
				For each frame %
				the figure shows:
				\textbf{(Left)}  Input \threeD meshes for the human (pink) and the object (yellow). 
				\textbf{(Right)} The color-coded body mesh to show our interaction-aware attention; 
				blue denotes body vertices that are far from the object (\ie, irrelevant for the specific interaction), and
				red denotes vertices that are near the object (\ie, very relevant).
	}
	\label{fig:distances}
\end{figure}

Here, we use vertex-to-vertex distances, but introduce a novel ``interaction-aware'' attention that focuses %
more on body vertices that are important for interaction (\eg, hands for grasping, feet for walking)
and less to irrelevant vertices (\eg, knees are less relevant than the hand for grasping). 
Our “interaction-aware” attention is formulated as:
\begin{equation}
     I_w(d) = \exp{\left(-w \times d\right)}, \quad  I_w : \reals^D -> \reals^D, \quad w>0
\label{eq:interaction_aware}
\end{equation}
where $d \in \reals^D$ is the distance vector and $w$ is a
learnable parameter. 
This gives exponentially more attention to vertices relevant for interaction.
This attention is visualized in \cref{fig:distances}; the attended body areas are meaningful. 
We set $w=5$, which empirically results in realistic grasps and motions.

\subsection{``Goal'' Network (\Gnet)}    \label{sec:Gnet}
\Gnet is a conditional variational auto-encoder (cVAE)\cite{kingmawelling2013} that generates a whole-body grasp, conditioned on the given object and its location. 
To do this, we first encode %
whole-body grasps into an embedding space.

\textbf{Input:}
The input $X$ to the encoder is:
\begin{equation}
X = 
\left[
\smplxparamssubject, \shapesubject, \vertssubject, \offsetssubject,
\headorient, \translobject, \basisobject
\right]
\label{eq:gnet_enc_inputs}
\end{equation}
where $\smplxparamssubject$ are the \smplX parameters,
$\vertssubject  \in \reals^{400\times3}$ are the \threeD coordinates of the sampled \smplX vertices,
$\headorient    \in \reals^{3}$          is a unit vector for head orientation, 
$\translobject  \in \reals^{3} $         is the object translation and 
$\basisobject   \in \reals^{1024} $      is the Basis Point Set (BPS) \cite{BPS19} representation of the \threeD object shape. 

Let $\offsets(\verts^{\mathrm{s}}, \verts^{\mathrm{t}}) \in \reals^{N \times 3} $ be a function
that computes offset vectors from the vertices of the source mesh $\verts^{\mathrm{s}}$
to the closest vertices of the target mesh $\verts^{\mathrm{t}}$: 
$\offsets_i(\verts_i^{\mathrm{s}}, \verts_i^{\mathrm{t}}) = $
\begin{equation}
    I_w\left(\normmse{\verts_i^{\mathrm{s}}- \verts_k^{\mathrm{t}} }\right), 
    k = \argmin_j{\normmse{\verts^{\mathrm{s}}_i - \verts^{\mathrm{t}}_j }}. 
\end{equation}
Finally let $\offsetssubject = \offsets(\vertssubject, \vertsobject) $, \ie
the offset vectors from the sampled body vertices,
$\vertssubject$, to the closest object vertices, $\vertsobject$.

At training time, the encoder $\encoder^{\text{G}}$ maps the inputs $X$ to the
parameters of a normal distribution $\mu, \sigma \in \reals^{16}$.
We then sample a latent whole-body grasp code $\graspcode \in \reals^{16}$ from this distribution 
using the re-parameterization trick \cite{kingmawelling2013}. 
Note that during inference we use the 16-dimensional standard normal distribution to sample
$\graspcode \sim \mathcal{N}(0, I)$.
Afterwards, the decoder takes 
the grasp code $\graspcode$, the
input conditions for the object, $\bm{C} = [\basisobject, \translobject]$,
and predicts \smplX parameters $\smplxparams$,
the head direction vector $\hat{\headorient}$, and 
offset vectors $\hat{\offsets}^{\righthand}$ 
from $99$ sampled right-hand vertices, $\vertssubject_{\righthand} \subset \vertssubject$ to object vertices. 

Both the encoder and decoder use fully-connected layers with skip connections, 
\Gnet is trained end-to-end and the loss is defined as 
$\loss  = \lambda_{\verts}            \loss_{\verts}            + $
\begin{align}
\lambda_{\verts}^{\righthand} {\loss_{\verts}^{\righthand}} +
\lambda_p {\loss_p} + \lambda_h {\loss_h} + \lambda_{\offsets^{\righthand}} \loss_{{\offsets^{\righthand}}} + \lambda_{KL} {\loss_{KL}} 
\label{eq:gnet_train_loss}, %
\end{align}
where
$\loss_{\verts}              = \normabs{\vertssubject - \vertssubjectGT}$, 
$\loss_{\verts}^{\righthand} = \normabs{\vertsrh      - \vertsrhGT}$, 
$\loss_p                     = \normmse{\smplxparams  - \smplxparamsGT}$, 
$\loss_\head                 = \normmse{\headorient   - \headorientGT}$, 
$\loss_{\offsets}^{rh}          = \normabs{\offsets^{\righthand}    - \offsetsGT^{\righthand}}$, and
$\loss_{KL}$ denotes the Kullback-Leibler divergence. 
The hat denotes regressed quantities; the non-hat variables are ground truth. 
For the exact architecture of \Gnet, see \supmat

We make two empirical observations:
(1) Networks struggle to predict accurate \smplX parameters, possibly due to their non-Euclidean space. 
(2) Networks predict \change{interaction features} in a Euclidean space much more precisely. 
These observations are in line with recent work \cite{zhang2021mojo,zanfir2021thundr,Loper:SIGASIA:2014}, but we go beyond them in regressing \threeD \emph{offsets} together with \smplX parameters, instead of regressing point positions and fitting \smplX to these. 
We leverage offsets in an optimization step to refine our \smplX predictions.

\textbf{\Gnet Optimization:}
We leverage the predicted offsets to refine our \smplX predictions with optimization post processing. 
Specifically, 
we \emph{optimize} over \smplX pose, $\posesubject$, and translation, $\translsubject$, 
initialized with \Gnet's predictions. %
Instead of hand-crafted contact constraints \cite{wang2021synthesizingLongTerm,Corona_2020_CVPR,hasson_2019_obman} during optimization, 
we use data-driven constraints \emph{generated} from \Gnet. %
Specifically, we use: %
(1) hand-to-object vertex offsets, 
(2) head-orientation and 
(3) pose coupling to the initial value, and 
(4) foot-ground penetration. 

In technical terms, 
for refining the hands to realistically grasp the objects, 
we define a $l_1$ term between 
the offsets $\hat{\offsets}^h$ generated from the \Gnet, and 
offsets computed online from \smplX's hand vertices to the closest object vertices: %
\begin{equation}
\energy_{\offsets}(\posesubject; \translsubject; \hat{\offsets}^{\righthand}) = \normabs{
\offsets(\verts^{\righthand}( \posesubject; \translsubject, \vertsobject) - \hat{\offsets}^{\righthand} }     \text{.}
\label{eq:gnet_opt_vertices}
\end{equation}
Pose and translation coupling discourages deviations from the initial ones:

\begin{equation}
\energy_{\posesubject}(\posesubject; \hat{\posesubject})  =  \normmse{ \posesubject - \hat{\posesubject} }, \quad
\energy_{\translsubject}(\translsubject; \hat{\translsubject})  =  \normabs{ \translsubject - \hat{\translsubject} }
\text{.}
\label{eq:coupling_pose}
\end{equation}

Similarly, head-orientation coupling is formulated as:
\begin{equation}
E_\head(\posesubject; \translsubject; \headorientGT) = \normmse{ \headorient(\posesubject; \translsubject) - \headorientGT}   \text{.}
\label{eq:coupling_head}
\end{equation}
Finally, we find the lowest vertex of the body along the y-axis (vertical axis)
and enforce its y-coordinate to be zero to have contact
and prevent penetration using: %
\begin{equation}
    \energy_\foot = \verts^{\mathrm{s}}_{k,y}, \quad k = \argmin_j{\verts^{\mathrm{s}}_{j,y}
\text{.}}
\label{eq:foot_penetration}
\end{equation}
Our final energy is a combination of the above five terms: %
\begin{equation}
\bm{E} = \lambda_v \energy_d +\lambda_t \energy_t +  \lambda_p \energy_p + \lambda_h \energy_h + \lambda_f \energy_f \text{.}
\label{eq:gnet_opt_loss}
\end{equation}
\change{The efficacy of our optimization post processing using the predicted Euclidean-space \change{interaction features}, is evaluated in the next section with a perceptual study (\cref{tab:MNet_results_amt})}.

\begin{figure}
    \centering
    \includegraphics[trim=000mm 000mm 000mm 000mm, clip=true, width=1.00 \linewidth]{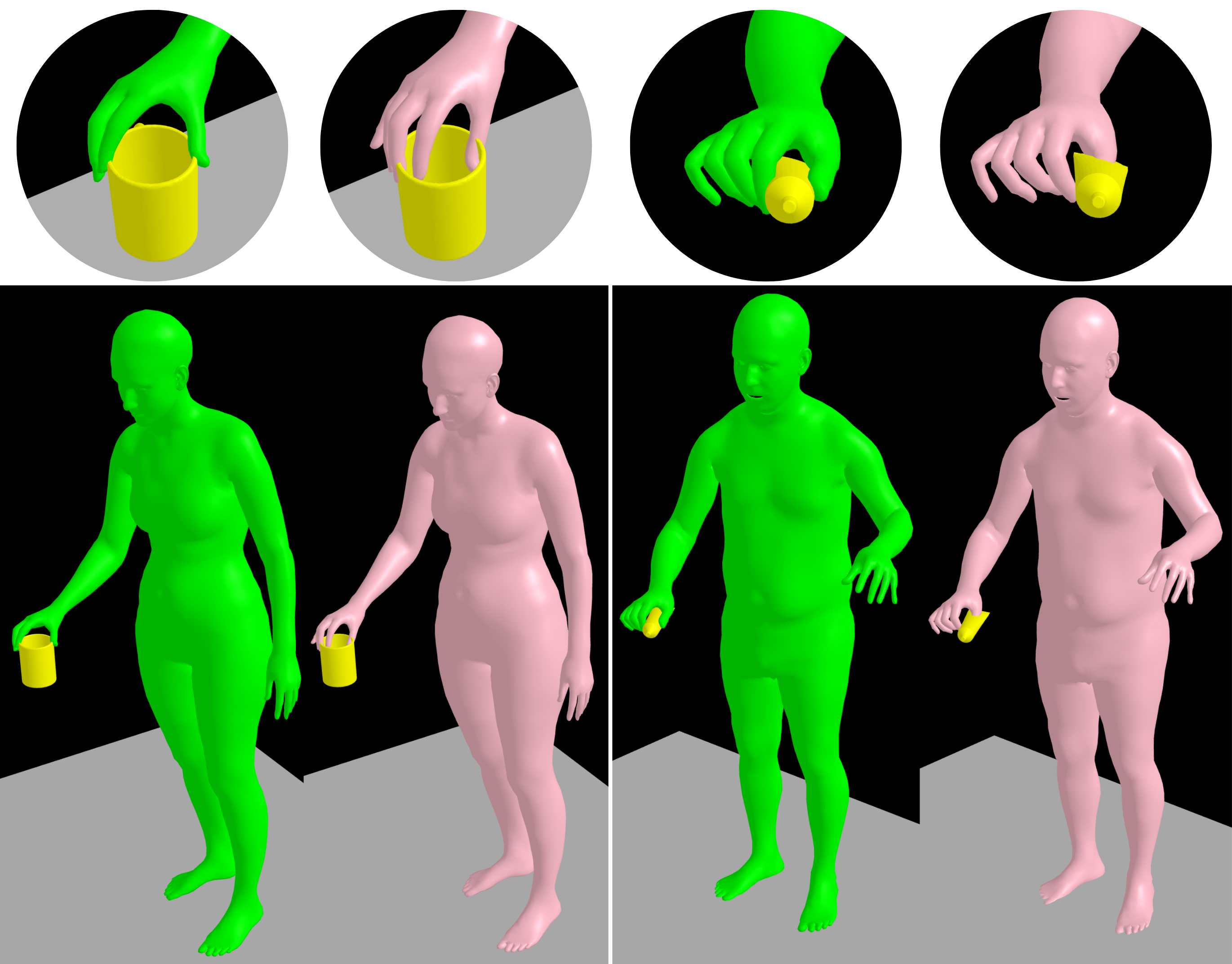}
    \caption{
            Generated \smplX grasp poses from \gnet (\cref{sec:Gnet}), 
            before (pink) and after optimization (green). 
            Results show that optimization-based post-processing effectively refines the initial prediction %
            towards a more realistic and physically plausible grasp. %
    }
    \label{fig:gnet_results}
\end{figure}

\subsection{Motion Network (\Mnet)}
\label{sec:Mnet}

\Mnet generates the motion from the starting to the ``goal'' frame; 
the latter is generated by \Gnet above. 
The length of a sequence depends on several factors, like the object location \wrt the body and the speed of motion. 
Therefore, to generate motion of arbitrary length, we use an auto-regressive network architecture \cite{hassan2021samp, starke2019neuralStateMachine}. 

\textbf{Input:}
\Mnet takes as input (auto-regressive fashion):
\begin{equation}
\bm{X}_p = [\smplxparamssubject_{t-5:t}, \shapesubject, \verts_t^{\subject}, \vertsvel_t^{\subject}, d_t^{\righthand}, b_\goal^{\righthand}]
\label{eq:mnet_inputs}
\end{equation}
\noindent where 
$\smplxparamssubject_{t-5:t}$   are \smplx parameters of the last $5$ frames, 
$\shapesubject$                 is the subject's shape, 
$\vertssubject_t$               and 
$\vertsvel_t^{\subject}$        are the locations and velocities of the sampled body vertices in the current frame, 
$d_t^{\righthand}$              are the hand %
                                vertex offsets from the current to the ``goal'' pose, and 
$b_\goal^{\righthand}$ is the \bps representation of the hand %
                                in the ``goal'' grasping frame. %
                                For this, we use the same \bps points as for the object. 
The \bps representation encodes the %
spatial relationship between the hand %
and the object in the ``goal'' frame, and is empirically important for ``guiding'' the motion towards a grasp with a good hand pose and hand-object contact. 
For our auto-regressive scheme, in agreement with \cite{starke2019neuralStateMachine}, we empirically find that using more than $1$ past frame leads to a smoother motion prediction; 
more than $5$ frames do not lead to noticeable improvement. 
 
Similar to \Gnet, along with predicting \smplX model parameters
$\smplxparams \in SE(3)$,
we also predict \change{interaction features} that lie in the Euclidean space. 
Empirically, this improves network inference, \ie the 
generated motion is smoother and better ``reaches'' the ``goal'' grasp. 
Unlike \cite{rempe2021humor}, where each motion frame depends only on $1$ past frame,
we %
find that \Mnet's generated motion quality
improves
as the number of future frames it generates grows; 
see \cref{tab:ablation_frames}. 

\textbf{Output:}
As output \Mnet network produces:
\begin{equation}
\bm{X}_f = [\Delta\theta_{t+10}^{\subject}, \Delta{t}_{t+10}^{\subject}, \Delta{v}_{t+10}^{\subject}, \Delta{d}_{t+10}^{\righthand}]
\label{eq:mnet_outputs}
\end{equation}
where $t+10$ denotes the future $10$ motion frames,
$\Delta\theta_{t+10}^{\subject}, \Delta{t}_{t+10}^{\subject}$, denote the change of \smplX pose and translation parameters, %
$\Delta{v}^{\subject}_{t+10}$       is the change of \smplX vertex locations, and 
$\Delta{d}^{\righthand}_{t+10}$     is the change of hand %
                                    vertex offsets.
All changes, $\Delta$, are relative to the current frame. 
In an auto-regressive fashion, \Mnet estimates \smplX parameters for ``motion'' poses, and then these 
are fed back to \Mnet as inputs (along with other ones) for the next iteration.
For the exact architecture of \Mnet, see \supmat

\begin{figure*}[t]
	\centering			%
	\includegraphics[trim=000mm 000mm 000mm 000mm, clip=true,width=1.0\linewidth]{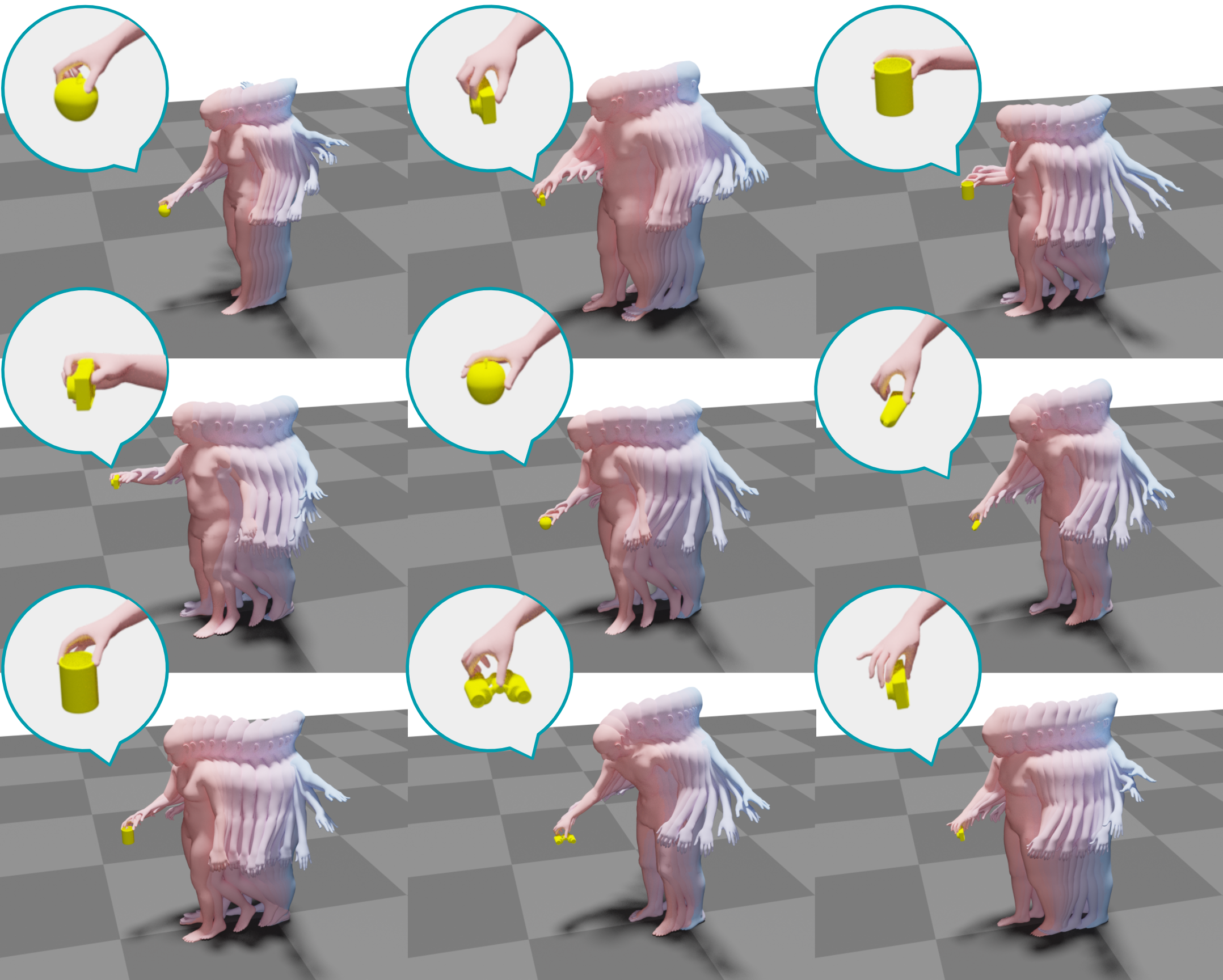}
    \caption{\modelname results: We show representative generated motions with different objects shapes, location as well as various body shapes.
    }
	\label{fig:results_goal}
\end{figure*}

\Mnet is trained end-to-end, with a loss similar to \Gnet. 
Specifically, we use a loss term on 
hand-to-object offsets, 
body parameters, body and hand vertices, similar to $\loss_{\offsets}^{\righthand}$, $\loss_p$, $\loss_{\verts}$, $\loss_{\verts}^{\righthand}$ in \cref{eq:gnet_train_loss} respectively. 

One common limitation of motion generation methods is ``skating'', \ie~foot sliding on the ground. 
To account for this, we define an additional loss term on foot vertices, when these are close to the ground. 
This loss, along with the computed input velocities for \cref{eq:mnet_inputs}, result in more realistic foot-ground contact; see \video in \supmat

\textbf{\Mnet Optimization:}
We refine \Mnet's generated motion with post processing based on optimization; 
this refines the motion for better ``reaching'' the ``goal'' grasping pose generated by \Gnet. 
Since we need precision only when the hand is very close to the object, 
we apply the optimization step only when \Mnet's estimated hand vertices get closer than $10$ cm to the ``goal'' hand vertex positions.

We follow \Gnet's scheme, and use \Mnet's predictions (\cref{eq:mnet_outputs}) as constraints, instead of hand-crafted ones. 
We first compute the average value of \Mnet's predicted hand-vertex velocities, $\dot{v_t}^{\righthand}$. 
Then, we linearly interpolate between the ``goal'', $v_{\goal}^{\righthand}$, and ``current'', $v_{t}^{\righthand}$, hand vertices: %
\begin{equation}
v_{\TODO{t+1}}^{\righthand} = v_{t}^{\righthand} + \lVert \dot{v_t}^{\righthand}  \rVert\times \hat{l},  \hat{l} =\frac{\vec{v_{\goal}}^{\righthand} - \vec{v_t}^{\righthand}}{\lVert \vec{v_{\goal}}^{\righthand} - \vec{v_t}^{\righthand} \rVert}
\label{eq:mnet_opt_linear_interp}
\end{equation}
where 
$\lVert \dot{v_t}^{\righthand} \rVert$      is the average-velocity magnitude,     and 
$\hat{l}$                                   is the (unit) vector pointing from ``current'' to the ``goal'' hand vertices. %
In practice, we ``force'' hands to move towards the ``goal'' grasp in a (locally) linear trajectory. %
Since our focus here is the hand grasp, for the rest of the body we keep the pose and velocity that \Mnet predicts.

The optimization objective function $L$ uses loss terms 
on        hand vertices,   $\loss_{\verts}^{\righthand}$, and 
on \smplX pose parameters, $\loss_p$, similar to the ones described for \cref{eq:gnet_train_loss}, and 
has the form:
\begin{equation}
L = \lambda_{p} L_p + \lambda_{\righthand} L_{\righthand}     \text{.}
\label{eq:mnet_opt_loss}
\end{equation}

\subsection{Implementation Details}

\textbf{Optimization details:}
For both \Gnet's and \Mnet's optimization-based post processing, we perform gradient descent with Adam \cite{adam} to optimize \smplX parameters. 

\textbf{Training data:}		\label{sec:method_Data}
For training both \Gnet and \Mnet, 
we use the \grab dataset \cite{GRAB:2020}, which contains whole-body \threeD~\smplX humans grasping \threeD objects. Please refer to \supmat for the details of data preparation.

\section{Experiments}
\label{sec:experiments}

\subsection{Qualitative Experiments}	\label{sec:experiments_Qualitative}

We show examples of \gnet's generated grasp before and after optimization in \cref{fig:gnet_results}.
Results show that \Gnet generates plausible body pose and head orientation for static grasps, but the hand grasps have room for improvement. %
The optimization step refines hand grasps so that they are more realistic and physically plausible. 
We show several representative motions generated by \Mnet with different objects, locations as well as various body shapes in \cref{fig:results_goal}.
For more results, please see our \video and \supmat

\subsection{Quantitative Experiments}
\label{sec:experiments_Quantitative}

\qheading{Perceptual Study:} To quantitatively evaluate
the generated results from \Gnet and \Mnet,
we perform a perceptual study through \ac{amt}. 

\qheading{\Gnet:}
For each test-set object, we use \Gnet to generate $2$ ``goal'' whole-body grasps. 
We render a ``turntable animation'' of 
the generated grasps, before and after optimization,
as well as the corresponding ground-truth grasps.
Participants are asked to rate the quality of $4$ features:
(1)     grasping pose, 
(2)     foot-ground contact, 
(3)     hand-object grasp, and
(4)     head orientation.
They rate the realism of each feature using a Likert scale of scores between $1$ (unrealistic) to $5$ (very realistic). 
Each grasp is evaluated by at least $10$ participants. 
To remove invalid ratings, \eg, participants that do not understand the task, we use catch trials similar to \grab \cite{GRAB:2020}.
The results of the evaluation are reported in \cref{tab:GNet_results_amt}. 
The study shows the effectiveness of the optimization step, especially on making the hand grasps more realistic.

The study shows that the optimized grasps have a better quality in grasping pose and head orientation compared to the ground truth. This is because in a subset of the \grab dataset the subject looks away while grasping the object, but in \gnet results the head is always oriented towards the object. The higher rating in the feet-ground penetration is due to the direct loss term in our optimization process which results in a better feet-ground contact. Overall, the quality of the generated grasps are close to the ground truth. 

\begin{table}
\centering
\renewcommand{\arraystretch}{1.2}
\scriptsize
\resizebox{1.00\linewidth}{!}{
\begin{tabular}{lccc}
\multicolumn{1}{l}{Metric} & \gnet & \gnet + Opt & Ground truth \cite{GRAB:2020}
\\
\toprule
Overall Grasping Pose $\uparrow$ & $3.89 \pm 0.93 $  & {$3.98 \pm 0.94 $}     & $3.78 \pm 1.06 $  \\
Foot-Ground Contact   $\uparrow$ & $3.98 \pm 1.06 $  & {$4.10 \pm 0.93 $}     & $3.82 \pm 1.11 $  \\ 
Hand-Object Grasp     $\uparrow$ & $2.70 \pm 1.37 $  & {$3.63 \pm 1.16 $}     & {$3.98 \pm 1.04 $}  \\ 
Head Orientation      $\uparrow$ & $3.83 \pm 1.01 $  & {$4.01 \pm 0.97 $}     & $3.84 \pm 1.07$   \\
Average               $\uparrow$ & $3.60 \pm 1.22 $  & {$3.93 \pm 1.02 $}     & $3.86 \pm 1.07$   \\
\bottomrule
\end{tabular}
} 
\vspace{-0.5 em}
\caption{%
    Evaluation of \gnet results, without and with optimization post processing.
    We ask the study participants to rate the realism of the grasp from $1$ (unrealistic) to $5$ (very realistic).
    We report the mean rating value $\pm$ the standard deviation, computed across all valid study participants.
    Optimization post processing (``\gnet + Opt'') 
    improves all of the four studied features.
}
\label{tab:GNet_results_amt}
\end{table}

\qheading{\Mnet:}
We use \Mnet to generate grasping motions on the test set.
In a perceptual study we show participants generated sequences and ground-truth ones,
and ask them to rate: 
(1)     the overall body motion quality,
(2)     foot-ground contact and sliding,
(3)     hand-object grasp at the end of the motion,
(4)     and head orientation. 
\Cref{tab:MNet_results_amt}
shows that \modelname generates realistic grasping motions, that approach the realism of ground truth. 
Note that \Mnet has a harder task than \Gnet, as it generates a full motion instead of a static pose. 
\change{By comparing \cref{tab:MNet_results_amt,tab:GNet_results_amt}, note that ground truth is rated higher for motions than for static poses and this is harder for \Mnet to match, though scores are not much lower.}

\qheading{Foot-Sliding Metric:}
We evaluate the physical plausibility of the generated motion using a ``foot-sliding'' metric. %
For each sequence, both generated and ground-truth ones, we find the closest \change{vertex} of the body to the ground and measure \change{its} velocity. 
We consider a frame to contain a ``sliding'' \change{foot} if the change in location of the selected \change{foot vertex} is higher than $1 cm$ per frame.
The percentage of ``foot-sliding'' frames in the ground truth and in the \modelname-generated sequences is $6.7 \%$,
and $13.7 \%$ respectively.
\change{Although there is room for improvement, empirically \Gnet's motions have less sliding than existing work (on other data).}

\begin{table}

\centering
\renewcommand{\arraystretch}{1.2}
\small
\resizebox{0.85\linewidth}{!}{
\begin{tabular}{lcc}
\multicolumn{1}{l}{Metric} & \modelname & Ground-truth \cite{GRAB:2020}  \\ 
\toprule
Overall Body Motion      $\uparrow$  & $3.74 \pm 0.97$  & $4.20 \pm 0.90$  \\ 
Foot-Ground Contact      $\uparrow$  & $3.88 \pm 1.14$  & $4.18 \pm 1.05$  \\ 
Final Hand-Object Grasp  $\uparrow$  & $3.66 \pm 1.05$  & $4.32 \pm 0.91$  \\ 
Head Orientation         $\uparrow$  & $3.86 \pm 1.03$  & $4.18 \pm 1.00$ \\ 
Average                  $\uparrow$  & $3.79 \pm 1.05$  & $4.22 \pm 0.97$ \\ 
\bottomrule
\end{tabular}
}
\vspace{-0.5 em}
\caption{%
        \mnet motion generation evaluation:
        We ask participants to rate the generated and ground-truth motion sequences on a Likert scale of 1 (unrealistic) to 5 (very realistic).
        The factors considered 
        are  overall body motion realism, feet-ground contact,
        final hand-object grasp and head orientation.
        }
\label{tab:MNet_results_amt}
\end{table}

\subsection{Ablation Study} \label{sec:experiments_AblationStudy}

\qheading{Number of Output Frames:} Here we study the effect of the number of output motion frames in \mnet. 
To study this, we trained $5$ networks with different numbers of output frames ranging from $1$ to $10$. 
We report the comparison in terms of the pose errors of the \smplx model, vertex-to-vertex distance for the body, feet, and hands between the generated results and the ground-truth. 
Results in \cref{tab:ablation_frames} show that generating more frames in each iteration of the auto-regressive network helps generate more accurate results. 
Additionally, we have observed qualitatively that, for networks with a lower number of frames as output, sometimes the motion does not converge to a final grasp, and the hands gradually deviate from the object.

\begin{table}[!t]
\centering
\small
\renewcommand{\arraystretch}{1.2}
\resizebox{1.0\linewidth}{!}{
\begin{tabular}{lllllll}
\toprule
     Number &\vtov-Body $\downarrow$ &\vtov-Hand $\downarrow$  & \vtov-Feet $\downarrow$ & Pose  $\downarrow$   & Trans $\downarrow$        &  \\
      of output frames  &  &  &  &  &   &  \\
     \midrule
1   & 14.70         & 10.60         & 17.10         & 3.67          & 5.84          &  \\ 
2   & 11.40         & 7.75          & 12.4          & 3.77          & 4.29          &  \\ 
3   & 10.43         & 6.7           & 11.60         & 4.00          & 4.07          &  \\ 
5   & 9.66          & 5.58          & 9.40          & 4.00          & 3.42          &  \\ 
10 & \textbf{9.34} & \textbf{3.89} & \textbf{8.34} & \textbf{3.67} & \textbf{3.02} &  \\ \bottomrule
\end{tabular}
}
\vspace{-0.5 em}
    \caption{
        Comparison between several \mnet architectures with different number of motion frames as output. The ``v2v'' notation shows the vertex-to-vertex distance error in the reconstruction loss of each network. ``Hand'' represents the right hand, and  Pose and Trans are \smplx model parameters. The results clearly show the improvement of losses by increasing number of output frames.
    }
\label{tab:ablation_frames}
\vspace{-0.5em}
\end{table}

\section{Conclusion and Future Work}        \label{sec:discussion}

We introduce GOAL, the first model to generate realistic human motions to grasp previously unseen {3D} objects. We use two novel networks (\gnet and \mnet) to first generate a static ``goal'' grasp and then inpaint the motion between the frames. We exploit the ability of both networks to infer interaction features
in
Euclidean space and introduce an optimization step after each network
to improve the quality of the grasps and motion based on the regressed features.
The evaluation
shows that our framework is able to synthesize natural and
physically plausible grasping motions.

\modelname opens up many possibilities for future studies on grasping motion generation. Even though \modelname generates realistic grasping motions, it is constrained to be in a close distance to the object and can not generate motions when the body is far from the object. Future work should  extend this to synthesize longer walking motions, prior to interaction with objects. In addition, in this work we focus on human-object interaction; in future work we would like to combine \modelname with human-scene interaction models to generate scene-aware grasping motions.

\vfill

\qheading{Social Impact:}
While realistic motion generation has mostly positive use cases in VR/AR, games, and movies, with the recent advances %
in neural rendering and deepfakes, we see a possibility that our results could be used for full-body deepfakes. 
Being aware of this, we will make our models available only for research purposes.

\clearpage
{
\smallskip
\noindent
\textbf{\emph{Acknowledgements:}}
This research was partially supported by the International Max Planck Research School for Intelligent Systems (IMPRS-IS), the Max Planck ETH Center for Learning Systems (CLS), and the German Federal Ministry of Education and Research (BMBF): Tubingen AI Center, FKZ: 01IS18039B.
We thank Tsvetelina Alexiadis for the Mechanical Turk experiments, and Taylor McConnell for the voice recordings. \\
\textbf{\emph{Disclosure:}}
MJB has received research gift funds from Adobe, Intel, Nvidia, Facebook, and Amazon. While MJB is a part-time employee of Amazon, his research was performed solely at, and funded solely by, Max Planck. MJB has financial interests in Amazon, Datagen Technologies, and Meshcapade GmbH.
}

{\small
\bibliographystyle{config/ieee_fullname}
\bibliography{config/BIB}
\balance
}

\clearpage

\renewcommand{\thefigure}{R.\arabic{figure}}
\renewcommand{\thetable}{R.\arabic{table}}

\newcommand{\supmattitle}[1]{
  \twocolumn[
  \begin{center}
    {\large \bf #1\par}
    \vspace*{12pt}
  \end{center}
  ]
}

\twocolumn[
{
    \renewcommand\twocolumn[1][]{#1}
    \supmattitle{\ourTitleSUPMAT}
    \centering
    \begin{minipage}{1.00\textwidth}
    \centering			%
	\includegraphics[width=1.00 \linewidth]{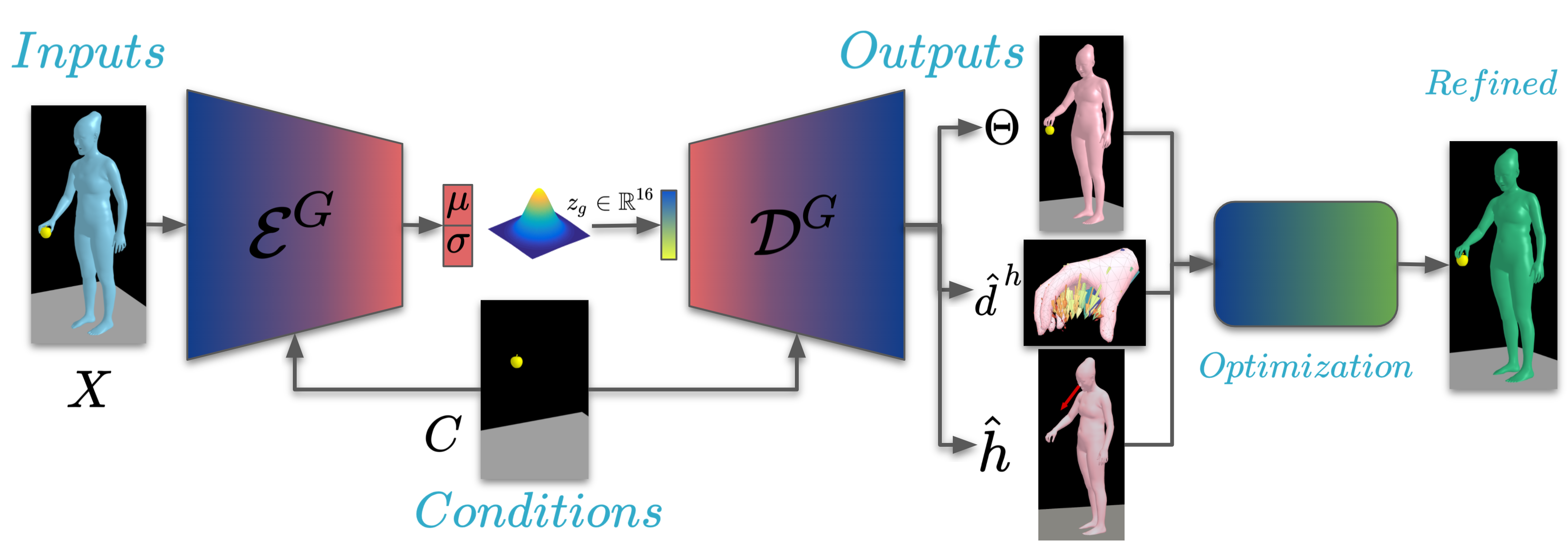}
    \end{minipage}
    \captionof{figure}{
            Architectural overview of the \Gnet network, as well as the optimization post-processing step (right-most part).
    }
    \label{fig:architecture_Gnet}
    \vspace*{+04.80em}
}
]

The supplemental material includes this document and a video.
Since the results involve movement, the video is important for evaluating the realism of our output.

\section{Data Preparation}				\label{sec:method}

\textbf{\Gnet data preparation:}
\Gnet generates static grasps. 
Therefore, from the \grab dataset, we collect all frames with right-hand grasps, for which participants grasp the object in a stable way. 
For this, we follow the selection criteria used for \grabnet's~[\textcolor{green}{55}] %
training data. 
We then center the object at the origin along the horizontal plane, \ie, while preserving its height. 
In total, we collect $160K$, $26K$, and $12.5K$ frames for the training, testing, and validation set, respectively. 

\textbf{\Mnet data preparation:}
\Mnet generates motion. 
Therefore, from each sequence of \grab, we gather all frames from the starting one up to the frame where the right hand first establishes a stable grasp.  
For this, we use the same selection criteria as above for \Gnet. 
We then create several sub-sequences by sliding a $21$-frame long window over each sequence with a stride of $1$ frame. 
For each sub-sequence, we consider the first $10$ frames as ``past'' motion, the last $10$ frames as ``future'' motion, and the middle one as the ``current'' frame. 
Then, following [\textcolor{green}{54}], %
we make all ``past'' and ``future'' frames relative to the body coordinate system of the ``current'' frame, while keeping the gravity direction always upward. 
In total, we collect roughly $40K$, $7K$, and $3K$ motion sub-sequences for the training, testing, and validation sets, respectively.

\section{\Gnet Architecture}

For an architectural overview of \Gnet and its optimization-based post processing, see \ref{fig:architecture_Gnet}. 

\section{\Mnet Architecture}

\begin{figure*}
    \centering
    \includegraphics[width=1.0\textwidth]{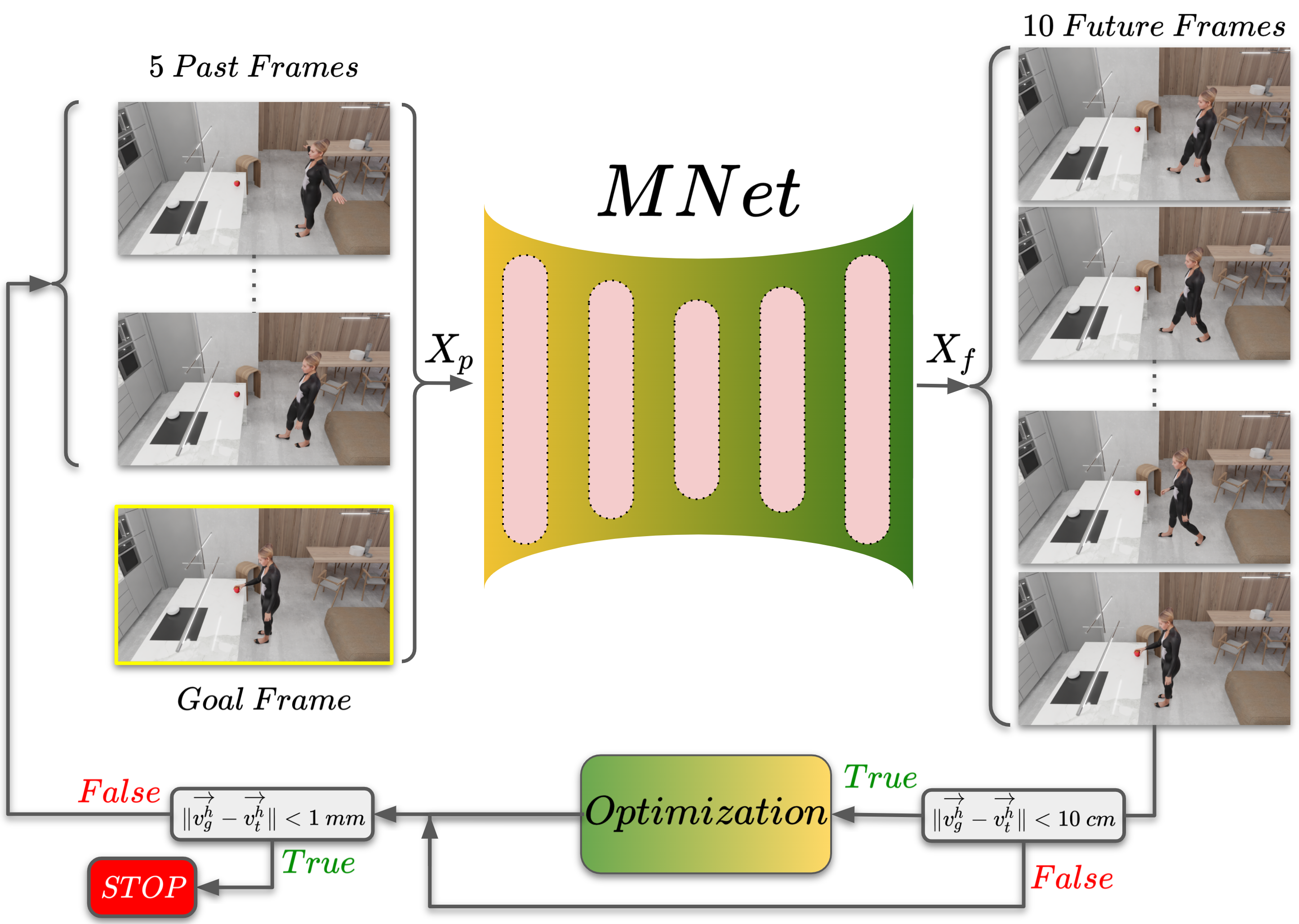}
    \caption{
            Architectural overview
            of the \mnet network, %
            as well as the 
            optimization post-processing step (bottom part).
            }
    \label{fig:architecture_Mnet}
\end{figure*}

For an architectural overview of \Mnet and its optimization-based post processing, see \ref{fig:architecture_Mnet}. 

\section{Video}

We provide a narrated video: 
(1)     explains our motivation, 
(2)     explains out method, and 
(3)     shows many results, including qualitative motion results.

\end{document}